\documentclass[runningheads]{llncs}
\usepackage[T1]{fontenc}
\usepackage{graphicx}
\usepackage{amsfonts}
\usepackage{amsmath}
\usepackage{multirow}
\usepackage{tabularx}
\usepackage{array}
\usepackage{caption}
\begin{document}
\title{Self-Supervised Pre-training Tasks for an fMRI Time-series Transformer in Autism Detection}

%
%


\author{Yinchi Zhou \inst{1} \and
Peiyu Duan \inst{1} \and
Yuexi Du \inst{1} \and
Nicha C. Dvornek\inst{1,2}}
%
\authorrunning{Zhou et al.}
%
\institute{Department of Biomedical Engineering \and
Department of Radiology and Biomedical Imaging, \\   
Yale University, New Haven, CT, USA}

\maketitle              
\begin{abstract}
Autism Spectrum Disorder (ASD) is a neurodevelopmental condition that encompasses a wide variety of symptoms and degrees of impairment, which makes the diagnosis and treatment challenging. Functional magnetic resonance imaging (fMRI) has been extensively used to study brain activity in ASD, and machine learning methods have been applied to analyze resting state fMRI (rs-fMRI) data. However, fewer studies have explored the recent transformer-based models on rs-fMRI data. Given the superiority of transformer models in capturing long-range dependencies in sequence data, we have developed a transformer-based self-supervised framework that directly analyzes time-series fMRI data without computing functional connectivity. To address over-fitting in small datasets and enhance the model performance, we propose self-supervised pre-training tasks to reconstruct the randomly masked fMRI time-series data, investigating the effects of various masking strategies. We then fine-tune the model for the ASD classification task and evaluate it using two public datasets and five-fold cross-validation with different amounts of training data. The experiments show that randomly masking entire ROIs gives better model performance than randomly masking time points in the pre-training step, resulting in an average improvement of 10.8\% for AUC and 9.3\% for subject accuracy compared with the transformer model trained from scratch across different levels of training data availability. Our code is available on GitHub \footnote{Code available at \url{https://github.com/ycaris/Self-Supervised\_fMRI}}.

\keywords{Autism \and fMRI \and Transformer \and Self-supervised Learning}
\end{abstract}
\section{Introduction}
Autism Spectrum Disorder (ASD) is a neurodevelopmental condition characterized by intellectual disabilities, impaired social interactions, language impairments, and repetitive behaviors. ASD affects a significant portion of the global population, with an estimated prevalence of approximately 1 in 100 children \cite{zeidan2022global}. However, ASD diagnosis is challenging due to the wide range of symptoms and severity, with current diagnostic practices heavily reliant on behavioral and developmental assessments that may be subject to the observer. In addition, the underlying causes of ASD are still unknown. 
To better characterize ASD phenotypes, functional magnetic resonance imaging (fMRI) has been used to investigate the brain activity of individuals with ASD. fMRI allows for the noninvasive measurement of brain signals through the recording of hemodynamic changes caused by neuronal activity. It provides high spatial resolution and can help locate brain functional activation areas, thereby mapping the connectivity patterns of the brain. Analysis of these signals could enable the identification of biomarkers, early diagnosis, and personalized treatment for ASD. 

Prior studies have shown promise in using the whole-brain functional connectivity calculated from the average of the time series of the regions of interest (ROIs) from resting-state fMRI (rs-fMRI) for the characterization and classification of ASD using machine learning \cite{van2010intrinsic,plitt2015functional,chen2015diagnostic,heinsfeld2018identification}. 
Recently, transformer models \cite{vaswani2017attention} have been applied to time-series rs-fMRI to model the dependency among different brain regions across time \cite{li2022transformer,bannadabhavi2023community}. Transformers are widely used in long sequence language tasks because the attention mechanism is capable of capturing global relationships between distant inputs \cite{devlin2018bert}. In ASD classification applications, Bannadabhavi et al. proposed a hierarchical transformer by learning the relationship of intra- and inter-community among brain regions \cite{bannadabhavi2023community}, and Li et al. constructed the positional encoding of the transformer-based model based on the functional connectivity matrix \cite{li2022transformer}. However, these approaches require the derived functional connectivity matrix as input to the model, rather than leveraging the original time-series data for analysis using the transformer model. 

Another challenge of adopting transformer models is that they often require learning on large training data to be successful. The model can first be pre-trained on a large unlabeled dataset, then finetuned on the specific task, improving data efficiency when the labeled data is limited for the target task, as is often the case for fMRI analysis \cite{ortega2023brainlm,malkiel2022self,malkiel2021pre}. 
The pre-training task usually involves reconstruction of the signal, either in an autoencoder style without masking \cite{malkiel2021pre,malkiel2022self} or with random masking of the input \cite{ortega2023brainlm}, akin to masked language modeling \cite{devlin2018bert}. 
The importance of the design of the pre-training masking strategy for temporal data has been noted in recent work~\cite{tong2022videomae}. 
Considering that the fMRI ROI time series is not only correlated within each individual ROI signal over time, but also between different ROIs as specified by the connectivity of different brain networks, a different masking strategy may be the key to pre-training.

In this work,  we investigated the use of different self-supervised pre-training tasks for an fMRI ROI time-series transformer model and evaluated their effects on data efficiency in learning the downstream fine-tuning ASD classification task. 
We evaluated our method on the public ABIDE \cite{di2014autism} and ACE datasets using five-fold cross-validation. The experiments show the effectiveness of transformer pre-training compared with the scratch transformer model even when trained with 100\% of the available data, with the pre-training strategy of randomly masking whole ROIs giving better model performance in the downstream ASD classification task than randomly masking time points in the pre-training step.

\section{Methods}
\subsection{Self-Supervised Training Framework Overview}
\begin{figure}[t]
\includegraphics[width=\textwidth]{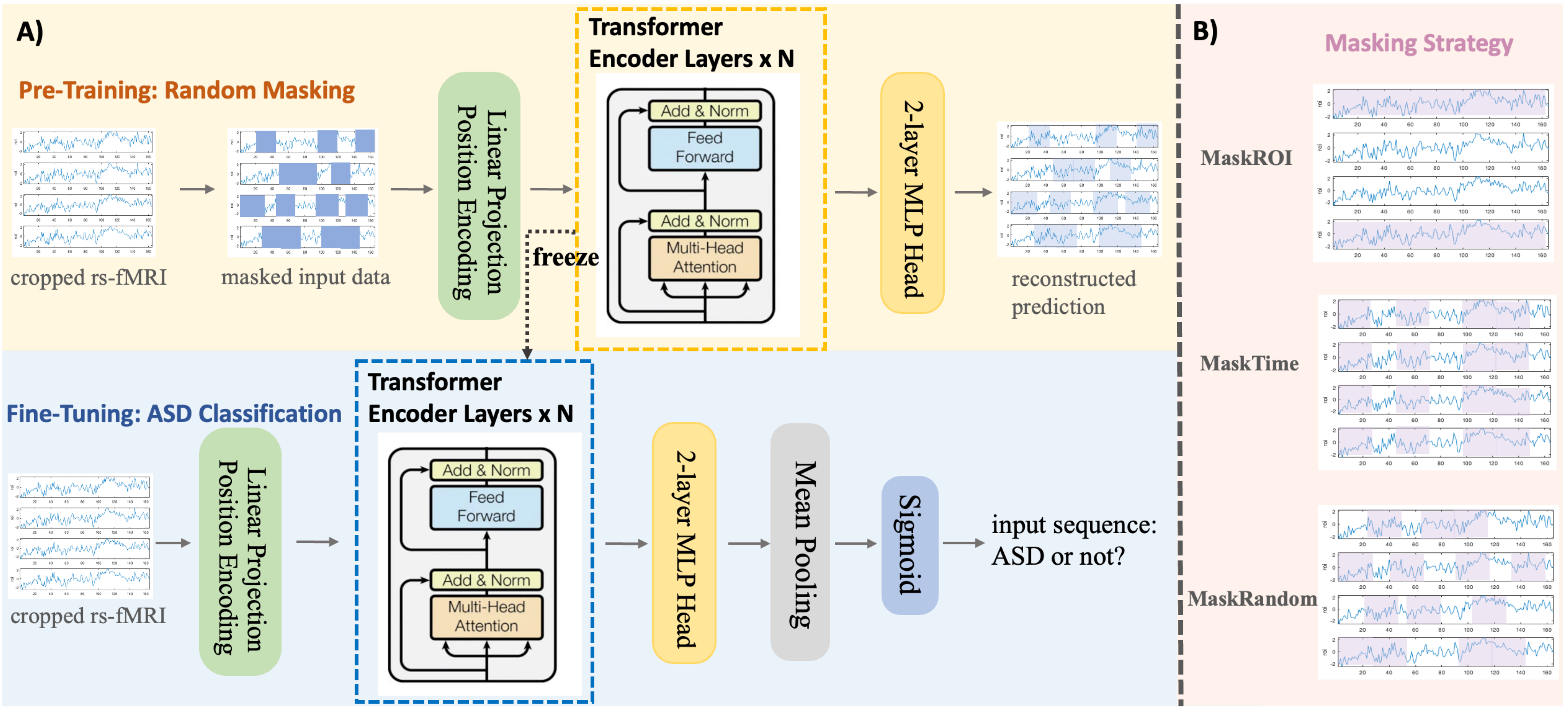}
\caption{Framework overview. A) The proposed self-supervised training workflow consists of a pre-training stage and a fine-tuning stage. The cropped rs-fMRI shown is obtained after data augmentation. B) Three masking strategies that are used in the pre-training tasks. Each row is the time-series data for one ROI.} \label{fig1}
\end{figure}

Our self-supervised training framework consists of two stages, the pre-training phase to reconstruct randomly masked time-series fMRI sequences and the fine-tuning phase to train a classifier on top of the transformer encoder for ASD classification (Fig.~\ref{fig1}). During the fine-tuning stage, we freeze the parameters of the transformer encoder layers and train an additional two-layer multilayer perceptron (MLP) classifier head to perform the ASD classification. 

\subsection{Model Architecture}
The input to the model is an fMRI time-series of length $T$ from $R$ ROIs. A linear projection layer maps this data to the hidden dimension, and sinusoidal positional encoding creates the embedding for the projected data. The transformer-based model has $N$ transformer encoder layers to extract features from time-series fMRI. The transformer encoder consists of a multi-headed self-attention module, position feed-forward network, residual connectivity, and layer normalization. The self-attention mechanism is the core component of the transformer model, which calculates the importance of individual tokens with respect to the input sequence. From the input tokens sequence $X$, we compute three embeddings: query ($Q$), key ($K$), and value ($V$). These embeddings are obtained with the formula below where $W_Q, W_K, W_V$ represent the learned weight matrices.
\begin{equation}
    Q = XW_Q,\quad K = XW_K,\quad V = XW_V
\end{equation}
The scaled attention is calculated from the embedding above to determine the importance of each token: 
\begin{equation}
    Attention (Q,K,V) = softmax(\frac{QK^T}{\sqrt{d_k}})V
\end{equation}
In our model, we use a multi-headed self-attention where we split Q, K, and V to $h$ sub-embeddings and project into $d_k$, $d_k$, $d_v$ dimensions, and $d_k = d_v = d_m/h$, where $d_m$ is the output dimension of the transformer encoder layers:
\begin{align}
    &MultiHead (Q,K,V) = Concat(head_1,...,head_h)W^O \\
    &head_i = Attention(QW^Q_i,KW^K_i,VW^V_i)
\end{align}
given the dimension $W_i^Q\in\mathbb{R}^{d_{m}\times d_k}$, $W_i^K\in\mathbb{R}^{d_{m}\times d_k}$, $W_i^V\in\mathbb{R}^{d_{m}\times d_v}$, $W^O\in\mathbb{R}^{hd_v\times d_{m}}$
Following the transformer encoder layers, a two-layer MLP head is used to reconstruct masked fMRI sequences, and the reconstruction loss is calculated from the cropped sequences and predicted sequences. In the fine-tuning stage, we re-use the same transformer encoder layers to encode the projected inputs, and then train a new MLP head for the downstream task. The output is passed into a sigmoid layer for final classification as follows:
\begin{equation}
    Probability = Sigmoid(AvgPool(MLP(ReLU(MLP(Z)))))
\end{equation}
where $Z$ represents the output from the transformer encoder layer.

\subsection{Random Masking in the Pre-training Stage} \label{masking_methods}
To investigate the effectiveness of different pre-training tasks, we construct three types of random masks (Fig.~\ref{fig1}B). 1) MaskROI: randomly select ROIs and mask the entire time periods of the rs-fMRI sequences. By masking whole ROIs, the model needs to reconstruct the missing indices from non-masked ROIs, enforcing the model to better learn the dependencies among the activity in different brain regions. 2) MaskTime: randomly select time points and mask all ROIs of those time points. This strategy is similar to that used in masked language modeling where random word tokens are masked during pre-training. The model needs to extract information from neighboring time points of specific ROIs. The resulting model may better learn the whole brain signal changes over time.  3) MaskRandom: first randomly select time points and then randomly select ROIs of those time points to mask. This strategy is similar to that used in prior self-supervised pre-training work in fMRI \cite{ortega2023brainlm}. The masking ratio is randomly set to 0.25 or 0.5. For each strategy, to create the mask, we set the value of the selected indices as zero.

\section{Experiments}

\subsection{Datasets and Pre-processing}
\subsubsection{ABIDE I} Autism Brain Imaging Data Exchange (ABIDE) I is a multi-site public dataset including 1112 subjects that were collected from 17 international sites \cite{di2014autism,craddock2013neuro} (age: 17.0 $\pm$ 8.0 years; 948 males and 164 females). It includes rs-fMRI images, T1 structural brain images and phenotypic information for each patient. The preprocesseed data using the Configurable Pipeline for the Analysis of Connectomes (CPAC) was downloaded from \cite{craddock2013neuro}. After the quality check, 886 subjects (409 with ASD, 477 healthy controls) were used for model training and testing. The mean time-series for each ROI was extracted using the AAL atlas which parcellated the brain into 116 ROIs \cite{tzourio2002automated}. The mean time-series from each ROI was standardized. 

\subsubsection{ACE} The Autism Centers of Excellence (ACE) public dataset\footnote{Data available from \url{https://nda.nih.gov/edit\_collection.html?id=2021}} includes comprehensive imaging, behavioral, and other data from a sex-balanced cohort of 526 ASD and neurotypical youth from 4 sites (ages: 13.3 $\pm$ 2.9 years). After quality control and filtering for missing data, 282 subjects (140 with ASD, 142 healthy controls) were used in the experiments, including 141 females and 141 males. The data were pre-processed using fMRIPrep, and mean ROI time-series were extracted using the AAL atlas and standardized. 


\subsection{Experimental Settings}

\subsubsection{Implementation Details}

To increase the effectiveness of pre-training and reduce the overfitting problem on a small dataset, we used a random cropping method as data augmentation to boost the number of training data \cite{dvornek2017identifying}. Specifically, we randomly cropped 10 sequences with length of $T$=64 time points from the original time-series fMRI data for each subject and re-cropped every epoch to ensure the randomness. 

The experiments were performed in PyTorch on an NVIDIA RTX A5000 GPU. In the pre-training stage, the transformer encoder and MLP head was trained for 50000 steps with a batch size of 64, dropout rate of 0.1, weight decay of $10^{-5}$, and an initial learning rate of $10^{-4}$. We used the cosine learning rate scheduler and the AdamW optimizer. We used $N=6$ encoder layers with 8 heads for each layer. Mean squared error (MSE) loss between the reconstructed sequences and the cropped sequences was used for optimization. 
In the fine-tuning stage, the transformer encoder was frozen, and only the MLP classifier was trained using the inner fold data. The model was trained with a batch size of 16, dropout rate of 0.1, weight decay of $10^{-3}$, and an initial learning rate of $10^{-4}$. Binary cross-entropy loss was optimized for ASD classification. A transformer model for the ASD classification task was also trained from scratch as a baseline. 
A smaller transformer model with only 2 encoder layers and 4 heads was used when training from scratch to avoid the overfitting problem due to lack of data.
To predict the subject label, we applied a sliding inference window on the input sequences of 64 time points from the original time-series fMRI data for each subject and determined the class using the majority voting of input sequences.  

\subsubsection{Model Evaluation}
For model evaluation, we used a nested subject-wise 5-fold cross-validation, stratified by diagnosis labels. In each of the outer 5 folds, 80\% of the data was used for training, and 20\% of the data was used for testing. Note the outer fold test data was left out in both the pre-training and fine-tuning process to fairly evaluate the model performance on unseen data. The training data used in the pre-training stage was further split into the inner five folds to accommodate the variability of the fine-tuned model. In each inner fold, 80\% of the data from the inner fold was used for training and 20\% of the data was used for validation to select the best model parameters. The final model evaluation was performed on the left-out test data from five outer folds. In addition, to demonstrate that pre-training improves data efficiency on the downstream task, we used different amounts of training data for ASD classification on the ABIDE dataset in fine-tuning, including 20\%, 40\%, 60\%, 80\%, 100\% of the training data in each fold. 
For each of the five testing sets, we obtained a five-fold ensemble result by averaging the model outputs from the inner folds. To evaluate the performance of pre-trained models on out-of-domain data, we first performed pre-training using all of the ABIDE data and then fine-tuned it on the ACE dataset for ASD classification under subject-wise five-fold cross-validation.

Classification model performance was evaluated using receiver operating characteristic (ROC) curve analysis. Significant pairwise differences between models were assessed using paired two-tailed t-tests. For the ABIDE experiments, we used the results from the five cross-validation folds and conducted a two-way repeated measures ANOVA to analyze the effects of the two independent factors of pre-training strategy and the amount of available data during fine-tuning on classification performance. 
This statistical test aims to find the statistical significance attributable to not only each factor but also the interaction of the two factors. Statistical significance was assessed at the level of p < 0.05 for all tests.

\begin{figure}[t!]
\centering
\begin{minipage}{0.56\textwidth}
    \centering
    \includegraphics[width=\textwidth]{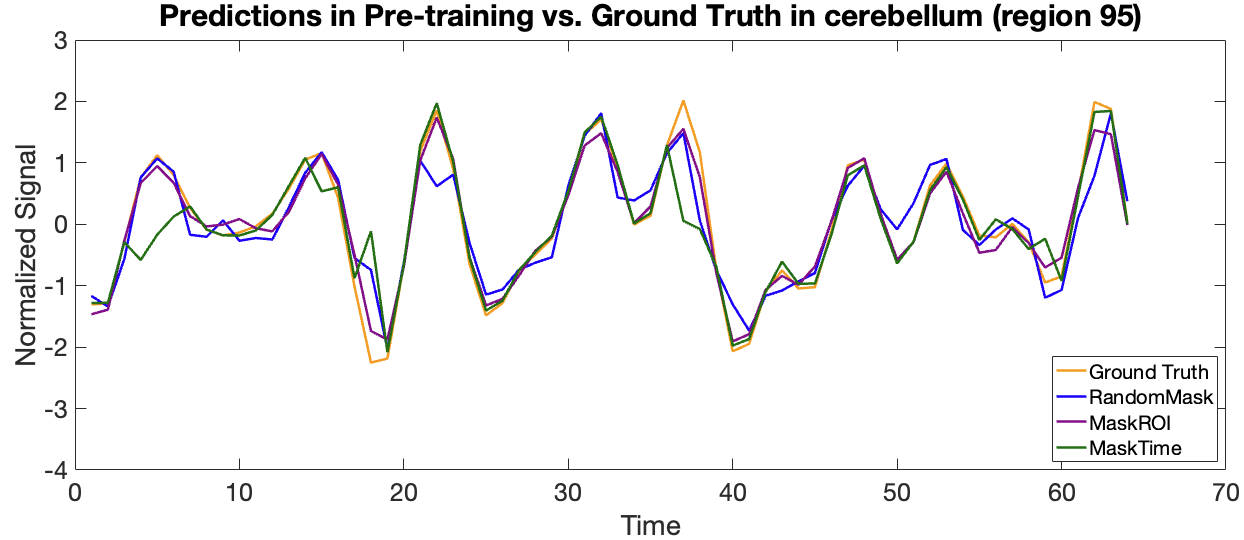}
    \caption{Visualization of example reconstructed sequences from the left-out testing data using different masking strategies.} 
    \label{fig4}
\end{minipage}\hfill
\begin{minipage}{0.41\textwidth}
    \centering
    \scriptsize
    \captionof{table}{MSE between reconstructed sequences and ground truth on the left-out testing data across five folds(mean $\pm$ std).}
    \begin{tabularx}{\textwidth}{|l|*{1}{>{\centering\arraybackslash}X|}}
        \hline
        \textbf{Masking Tasks} & \textbf{MSE} \\ \hline
        MaskRandom & \textbf{0.109 $\pm$ 0.002} \\ \hline
        MaskTime & 0.136 $\pm$ 0.002 \\ \hline
        MaskROI & 0.136 $\pm$ 0.047 \\ \hline
        
    \end{tabularx}
\end{minipage}
\end{figure}



\section{Results}
\subsection{Pre-training Results with Different Random Masking Strategies}
The pre-training task was to reconstruct the masked sequences with $T$=64 time points. The three different masking strategies were shown to be effective visually, where the reconstructed sequences have very similar patterns to the original data (Fig.~\ref{fig4}). MSE between the reconstructed sequences and the original data on the left-out testing data for ABIDE across five folds is shown in Table 1. Quantitatively, completely random masking had the lowest average MSE, and MaskROI had higher variability. While MaskRandom resulted in significantly lower MSE than MaskTime (p < 0.001), no significant difference was found between MaskROI and  MaskRandom or MaskTime (p > 0.05).

\begin{table}[t!]
\centering
\caption{Evaluation metrics for ASD classification on ABIDE with different pre-training strategies and different percentages of training data in the fine-tuning stage. The results are summarized from five testing sets. (mean $\pm$ std).}
\scriptsize
\begin{tabularx}{\textwidth}{|l|*{5}{>{\centering\arraybackslash}X|}}
\hline
\multirow{2}{*}{\textbf{Model}} & \multicolumn{5}{c|}{\textbf{AUC}} \\ \cline{2-6}
 & \textbf{20\%} & \textbf{40\%} & \textbf{60\%} & \textbf{80\%} & \textbf{100\%} \\ \hline
Scratch & 0.59 $\pm$ 0.05 & 0.61 $\pm$ 0.05 & 0.60 $\pm$ 0.05 & 0.62 $\pm$ 0.05 & 0.65 $\pm$ 0.07 \\ \hline
MaskRandom & 0.61 $\pm$ 0.07 & 0.66 $\pm$ 0.04 & 0.69 $\pm$ 0.06 & \textbf{0.71 $\pm$ 0.05} & 0.72 $\pm$ 0.05 \\ \hline
MaskTime & 0.57 $\pm$ 0.07 & 0.65 $\pm$ 0.05 & 0.69 $\pm$ 0.06 & 0.70 $\pm$ 0.04 & 0.72 $\pm$ 0.04 \\ \hline
MaskROI & \textbf{0.62 $\pm$ 0.07} & \textbf{0.67 $\pm$ 0.04} & \textbf{0.70 $\pm$ 0.04} & \textbf{0.71 $\pm$ 0.05} & \textbf{0.73 $\pm$ 0.04}\\ \hline
\multirow{2}{*}{\textbf{Model}} & \multicolumn{5}{c|}{\textbf{Subject Accuracy}} \\ \cline{2-6}
 & \textbf{20\%} & \textbf{40\%} & \textbf{60\%} & \textbf{80\%} & \textbf{100\%} \\ \hline
Scratch & 0.57 $\pm$ 0.06 & 0.57 $\pm$ 0.03 & 0.59 $\pm$ 0.05 & 0.58 $\pm$ 0.03 & 0.61 $\pm$ 0.05 \\ \hline
MaskRandom & \textbf{0.60 $\pm$ 0.04} & \textbf{0.62 $\pm$ 0.03} & 0.64 $\pm$ 0.04 & \textbf{0.66 $\pm$ 0.04} & \textbf{0.66 $\pm$ 0.04} \\ \hline
MaskTime & 0.57 $\pm$ 0.03 & 0.60 $\pm$ 0.02 & \textbf{0.65 $\pm$ 0.04 }& 0.65 $\pm$ 0.02 & 0.65 $\pm$ 0.02 \\ \hline
MaskROI & 0.59 $\pm$ 0.03 & 0.61 $\pm$ 0.02 & 0.64 $\pm$ 0.04 & \textbf{0.66 $\pm$ 0.06} & \textbf{0.66 $\pm$ 0.04}\\ \hline

\end{tabularx}
\label{abide_tab}
\end{table}

\begin{figure}[t]
\centering
\includegraphics[width=\textwidth]{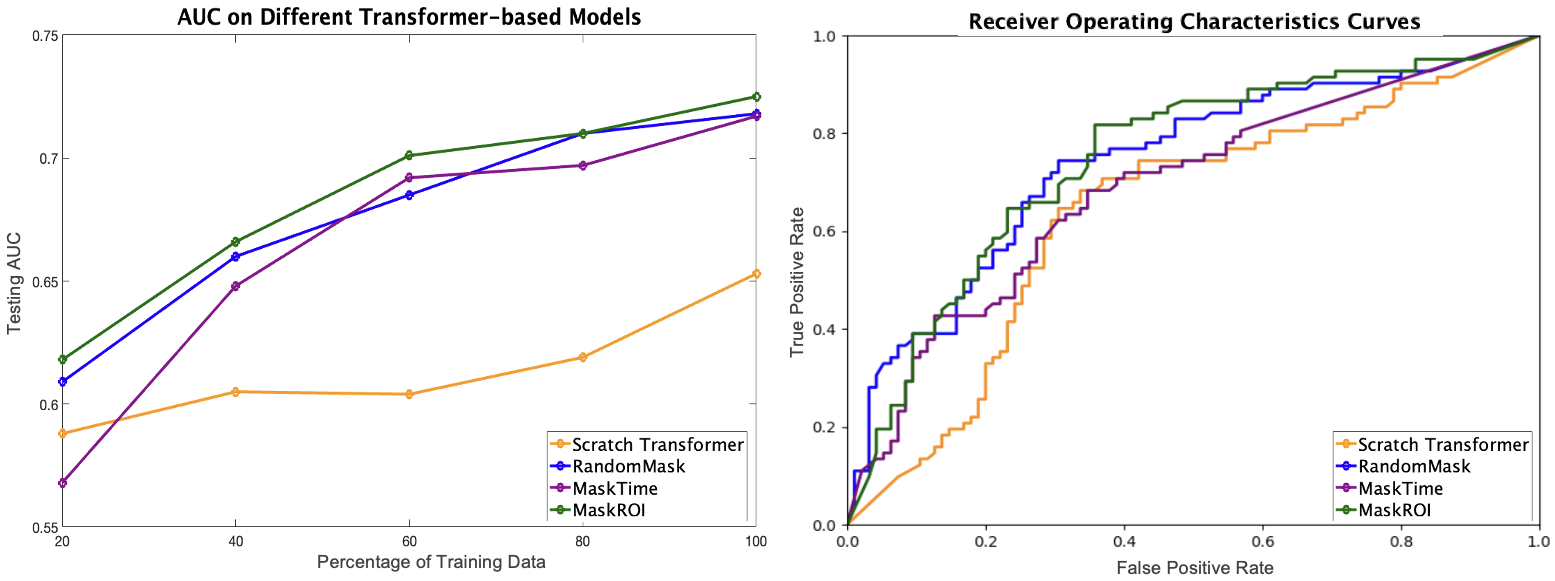}
\caption{Left: Testing AUC of ASD classification models learned from different percentages of training data using different pre-training masking strategies. Right: Example ROC curves of one test set using 100\% training data.} \label{abide_plots}
\end{figure}

\begin{table}[t]
\centering\
\footnotesize
\caption{Evaluation metrics for ASD classification on ACE dataset (mean $\pm$ std). Pre-trained models were learned on all ABIDE data, then fine-tuned on ACE.}
\begin{tabularx}{\textwidth}{|l|*{2}{>{\centering\arraybackslash}X|}}
\hline
\textbf{Model} & \textbf{AUC} & \textbf{Subject Accuracy} \\ \hline
Scratch Transformer & 0.56 $\pm$ 0.07 & 0.53 $\pm$ 0.04 \\ \hline
MaskRandom Finetune & 0.61 $\pm$ 0.07 & 0.58 $\pm$ 0.05 \\ \hline
MaskTime Finetune & 0.60 $\pm$ 0.05 & 0.56 $\pm$ 0.03 \\ \hline
MaskROI Finetune & \textbf{0.65 $\pm$ 0.04} & \textbf{0.61 $\pm$ 0.03} \\ \hline
\end{tabularx}
\label{ace_tab}
\end{table}

\subsection{Downstream ASD Classification Results}
The performance of ASD classification on the ABIDE dataset is shown in Table \ref{abide_tab} and Fig.~\ref{abide_plots}. Our results indicate that both AUC and subject accuracy improve as the percentage of training data increases for all models, which is likely to be attributed to the overfitting problem when limited training data is used. We also observe that all pre-training strategies resulted in fine-tuned models with higher average performance than the corresponding scratch model. The two-way repeated measures ANOVA for AUC indicated no significant interaction between the pre-training strategy and the amount of training data (p = 0.152), but showed significant effects for different pre-training tasks (p = 0.002) and the amount of training data (p = 0.023) alone.    
Specifically, a completely random mask yielded an average AUC improvement of 9.5\% compared to the scratch transformer model. Masking all ROIs at randomly selected time points (MaskTime) resulted in an average AUC improvement of 7.5\%, while masking all time points at randomly selected ROIs (MaskROI) led to an average AUC improvement of 10.8\%. 
Among different masking strategies, MaskROI consistently achieved the highest AUC and the highest or second-highest accuracy across all percentages. This may be attributed to the model's enhanced ability to learn the relationships between different brain regions, corresponding to better modeling of brain connectivity, which has shown to be widely affected in ASD patients \cite{kana2007inhibitory,rane2015connectivity,muller2011underconnected,hull2017resting}. Subject accuracy also improved with the use of the pre-trained model across all percentages. Differences in masking strategies were evident in sensitivity and specificity metrics (Supplementary Table S1). 
MaskROI exhibited the highest sensitivity yet lowest specificity, but also it was the only model that produced overall average sensitivity and specificity that were both greater than 0.5.

The performance of ASD classification on the ACE dataset is shown in Table \ref{ace_tab}.  Results showed that the fine-tuned models using all masking strategies outperformed the scratch transformer model for ASD classification, indicating the effectiveness of pre-training even with different datasets. The small size of the ACE dataset causes overfitting of the transformer model trained from scratch to the training data, leading to low AUC and subject accuracy on the test set. By using a pretrained encoder from the ABIDE dataset, we can train a classification head to mitigate the overfitting problem. Notably, the fine-tuned model with MaskROI achieved the highest AUC and accuracy and was the only pre-training method that resulted in a significantly higher performance than the scratch model (p = 0.006 for AUC, p = 0.040 for accuracy).  Furthermore, the MaskROI model performed significantly better than the MaskTime model (p = 0.048 for AUC, p = 0.020 for accuracy). These results highlight the importance of the choice of the pre-training task. 

\section{Conclusion}
In this work, we proposed a transformer-based self-supervised pre-training and fine-tuning framework for ASD classification using time-series fMRI data. The framework includes a pre-training stage where input sequences are masked using three different strategies: randomly masking time points, randomly masking ROIs, and completely random masking. We trained and evaluated our method on two datasets and observed significant improvements compared to the scratch transformer model without pre-training. Our results indicate that the performance differs among the three masking strategies used during pre-training. Randomly masking time points underperforms compared to the other two strategies while masking entire ROIs achieves the best performance on both the ABIDE and ACE datasets. Future work will focus on incorporating multimodal information into the current framework to further enhance classification accuracy.

\subsubsection{Acknowledgements}
This research is supported in part by the National Institute of Neurological Disorders and Stroke (NINDS) of the National Institutes of Health grant R01NS035193.  

\subsubsection{Disclosure of Interests}
The authors have no competing interests to declare that are relevant to the content of this article.

\bibliographystyle{splncs04}
\bibliography{Paper-0026}

\clearpage

\end{document}


\section{Supplementary Materials}

\makeatletter
\setcounter{figure}{0}
\setcounter{table}{0}
\renewcommand{\thefigure}{S\@arabic\c@figure}
\renewcommand{\thetable}{S\@arabic\c@table}
\makeatother

\subsection{Model Sensitivity and Specificity}
\begin{table}[h!]
\centering
\caption{Sensitivity and specificity for ASD classification on ABIDE with different pre-training stategies and different percentages of training data in the fine-tuning stage (mean $\pm$ std).}
\scriptsize
\begin{tabularx}{\textwidth}{|l|*{5}{>{\centering\arraybackslash}X|}}
\hline
\multirow{2}{*}{\textbf{Model}} & \multicolumn{5}{c|}{\textbf{Sensitivity}} \\ \cline{2-6}
 & \textbf{20\%} & \textbf{40\%} & \textbf{60\%} & \textbf{80\%} & \textbf{100\%} \\ \hline
Scratch & \textbf{0.52 $\pm$ 0.11} & \textbf{0.54 $\pm$ 0.25} & 0.40 $\pm$ 0.16 & 0.273 $\pm$ 0.22 & 0.46 $\pm$ 0.22 \\ \hline
MaskRandom & 0.33 $\pm$ 0.24 & 0.46 $\pm$ 0.09 & 0.49 $\pm$ 0.08 & 0.53 $\pm$ 0.10 & 0.52 $\pm$ 0.06 \\ \hline
MaskTime & 0.17 $\pm$ 0.13 & 0.38 $\pm$ 0.10 & 0.49 $\pm$ 0.16 & 0.45 $\pm$ 0.09 & 0.43 $\pm$ 0.04 \\ \hline
MaskROI & 0.37 $\pm$ 0.28 & 0.45 $\pm$ 0.14 & \textbf{0.50 $\pm$ 0.19} & \textbf{0.62 $\pm$ 0.15} & \textbf{0.59 $\pm$ 0.12} \\ \hline
\multirow{2}{*}{\textbf{Model}} & \multicolumn{5}{c|}{\textbf{Specificity}} \\ \cline{2-6}
 & \textbf{20\%} & \textbf{40\%} & \textbf{60\%} & \textbf{80\%} & \textbf{100\%} \\ \hline
Scratch & 0.62 $\pm$ 0.18 & 0.58 $\pm$ 0.25 & 0.75 $\pm$ 0.20 & 0.85 $\pm$ 0.16 & 0.74 $\pm$ 0.15 \\ \hline
MaskRandom & 0.82 $\pm$ 0.15 & 0.75 $\pm$ 0.11 & 0.77 $\pm$ 0.03 & 0.78 $\pm$ 0.07 & 0.79 $\pm$ 0.07 \\ \hline
MaskTime & \textbf{0.91 $\pm$ 0.08} & \textbf{0.80 $\pm$ 0.11} & \textbf{0.80 $\pm$ 0.05} & \textbf{0.82 $\pm$ 0.05} & \textbf{0.85 $\pm$ 0.02} \\ \hline
MaskROI & 0.78 $\pm$ 0.19 & 0.75 $\pm$ 0.11 & 0.79 $\pm$ 0.12 & 0.69 $\pm$ 0.12 & 0.72 $\pm$ 0.06\\ \hline
\end{tabularx}
\end{table}